\documentclass[10pt,twocolumn,letterpaper]{article}

\usepackage{iccv}              % To produce the CAMERA-READY version
\usepackage{multirow}
\usepackage{colortbl}  % For row coloring
\usepackage{xcolor}    % For text coloring
\definecolor{customgreen}{rgb}{0.0, 0.5, 0.0} % Adjust the RGB values to make the green darker
\usepackage{makecell} % For table
\usepackage{comment}
\usepackage{stfloats}

\definecolor{iccvblue}{rgb}{0.21,0.49,0.74}
\usepackage[pagebackref,breaklinks,colorlinks,allcolors=iccvblue]{hyperref}

\def\paperID{5069} % *** Enter the Paper ID here
\def\confName{ICCV}
\def\confYear{2025}

\title{Grounding-Aware Token Pruning: Recovering from Drastic Performance Drops in Visual Grounding Caused by Pruning}

\author{
Tzu-Chun Chien\textsuperscript{1}\quad
Chieh-Kai Lin \quad
Shiang-Feng Tsai\textsuperscript{1}\thanks{Equal contribution} \quad
Ruei-Chi Lai\textsuperscript{1}\footnotemark[1] \quad
Hung-Jen Chen\textsuperscript{1}\quad
Min Sun\textsuperscript{1}\\
\textsuperscript{1}National Tsing Hua University (NTHU)
}

\begin{document}
\maketitle
\begin{abstract}
Recent Multimodal Large Language Models (MLLMs) have demonstrated strong performance in visual grounding, establishing themselves as a general interface for various vision-language applications. This progress has driven the development of token pruning methods to mitigate the high computational costs associated with processing numerous visual tokens. However, we observe that pruning significantly weakens the model’s grounding ability, leading to incorrect predictions and drastic performance degradation. In Referring Expression Comprehension (REC), for instance, pruning causes the accuracy of LLaVA on the RefCOCO val set to drop from \textbf{56.14\%} to \textbf{15.34\%}. Our analysis identifies misaligned position IDs after pruning as the primary cause of this degradation, as both the order and value of these IDs are crucial for maintaining performance in grounding tasks. To address this issue, we propose Grounding-Aware Token Pruning (\textbf{GAP}), a simple yet effective adjustment to position IDs that recovers REC accuracy back to \textbf{51.42\%}, which is 90\% of the original performance in the without pruning setting, all while requiring no additional training, memory, or computational overhead. Applied to models such as Shikra, MiniGPTv2, and the LLaVA series, our method consistently improves performance across various token pruning strategies.
\end{abstract}    
\section{Introduction}
\label{sec:intro}

\begin{figure}[t]
    \centering
    \includegraphics[width=\columnwidth]{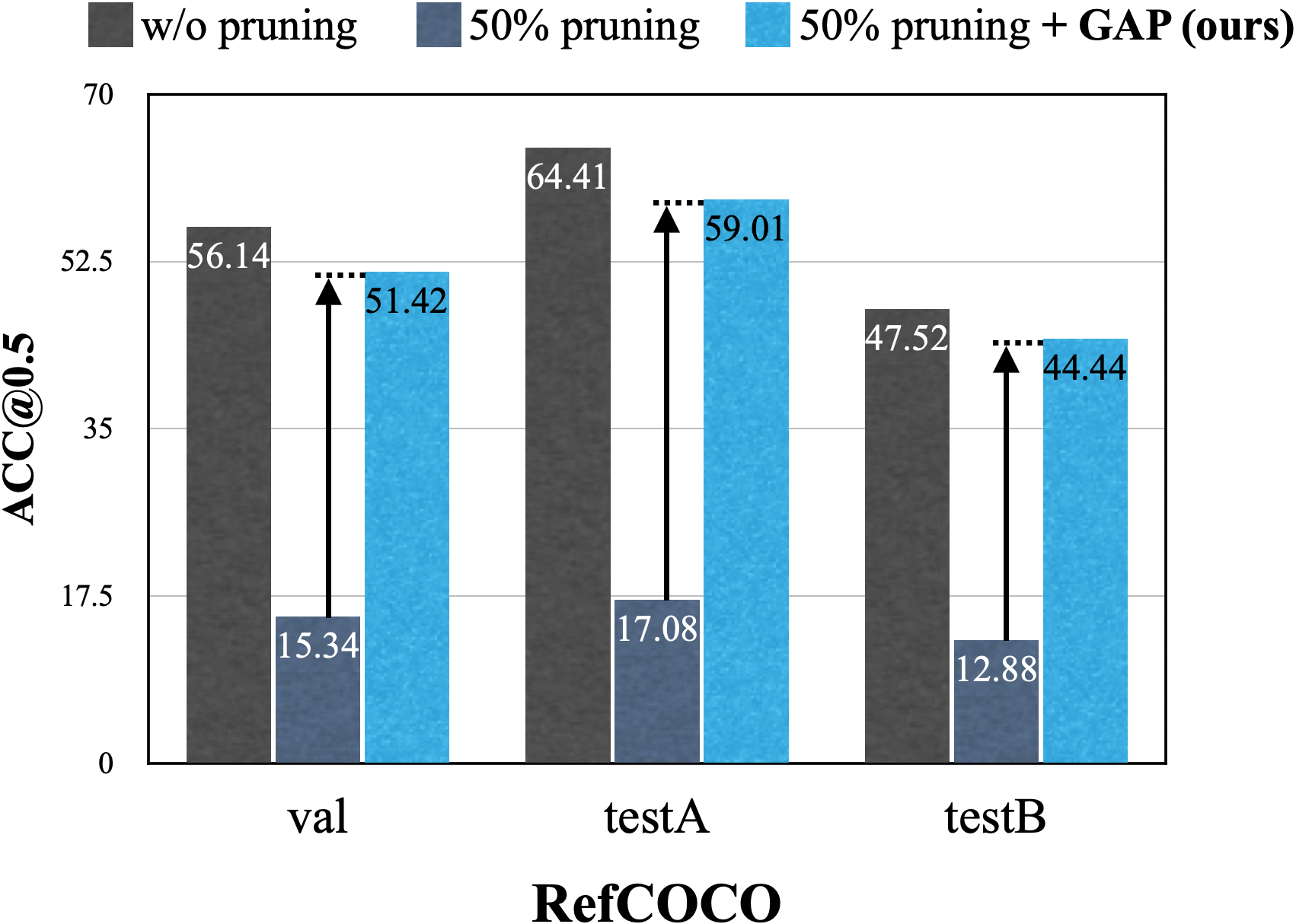}
    \caption{\textbf{Catastrophic performance drop.} Comparison of the performance of LLaVA on the visual grounding dataset RefCOCO. Performance drops drastically after pruning, but is restored after applying our approach, \textbf{GAP}, which preserves the grounding ability of MLLMs.}
    \label{fig:drastic_drop}
    \vspace{-0.5cm}
\end{figure}

Multimodal Large Language Models (MLLMs) have demonstrated impressive performance across a range of tasks, such as VQA \cite{gqa, vizwiz, okvqa} and image captioning. Notably, for Referring Expression Comprehension (REC) \cite{kazemzadeh-etal-2014-referitgame}, models like MiniGPTv2 and Shikra \cite{chen2023minigptv2, chen2023shikra} achieve results comparable to specialized models, underscoring their capacity for dense visual understanding, such as object localization and spatial reasoning.

As the number of visual tokens generated from high-resolution images continues to grow (e.g., 2k-8k tokens) \cite{liu2024llavanext, llavaonevision}, computational costs and inefficiencies increase. High memory usage, latency, and higher FLOPs hinder scalability. To address these issues, various token pruning methods \cite{prumerge, trim,cao-etal-2023-pumer, Zhang2024SparseVLMVT, Yu2024BalancingPA, Chen2024RecoverableCA} have been developed, which selectively reduce visual tokens while optimizing computational efficiency.

However, we observe that token pruning can degrade grounding ability, impairing the model’s understanding of spatial relationships. This results in significantly poorer bounding box predictions and a dramatic performance drop in grounding tasks that require fine-grained spatial information, such as REC, as shown in \Cref{fig:drastic_drop}.
Through investigation, we found that misaligned position IDs are the primary cause of performance drops after pruning. These IDs are crucial for maintaining the spatial coherence of visual tokens, and both the order and values of these IDs must remain intact to ensure optimal model performance. Based on this, we propose \textbf{G}rounding-\textbf{A}ware Token \textbf{P}runing \textbf{(GAP)}, a simple modification that corrects these misalignments and significantly improves REC performance. 

%For instance, in MiniGPTv2, accuracy is restored from 4\% to 80\% with random pruning, without the need for model retraining or additional computational overhead.

We applied GAP to six pruning methods, including, PruMerge \cite{prumerge}, TRIM \cite{trim}, and four baseline methods on LLaVA, demonstrating its ability to improve existing pruning strategies without additional overhead. To further validate our findings, we applied GAP to five models, including Shikra, MiniGPTv2, and the LLaVA series, confirming that this is a global issue affecting various training recipes and model architectures. These findings confirm that GAP consistently improves performance while preserving efficiency, making it a viable enhancement for grounding MLLMs.

Our contributions are summarized as follows:
\begin{itemize}
    \item We identify misaligned position IDs as the key factor behind grounding performance degradation in token-pruned MLLMs and propose \emph{GAP} to effectively recover lost performance.
    \item \emph{GAP} aligns position IDs without additional training, memory, or computational overhead, making it an efficient and practical solution.
    \item We validate \emph{GAP} across five models and six pruning methods, demonstrating its broad applicability and effectiveness in addressing this global issue.
\end{itemize}
% -----審稿

\section{Background and Related Work}
\label{sec:background}
\subsection{Multimodal Large Language Models (MLLMs)}
The expansion of large language models into multimodal versions has gained considerable attention in recent years. Multimodal Large Language Models (MLLMs) \cite{liu2023llava, liu2023improvedllava, liu2024llavanext, chen2023minigptv2, chen2023shikra} represent a significant advancement in artificial intelligence, extending the capabilities of traditional Large Language Models (LLMs) \cite{brown2020languagemodelsfewshotlearners, devlin2019bertpretrainingdeepbidirectional} and traditional Vision Language Models (VLMs) to process content across multiple modalities, inclusive of text, images, audio and video. 

Mainstream approaches in MLLMs use sequential visual representation, where images are encoded into vision tokens by a visual encoder \cite{openaiclip, eva, siglip} and sent into the LLM as the prefix content. With modal alignment pretraining and instruction fine-tuning \cite{liu2023llava, zhu2023minigpt}, modern MLLMs can tackle various tasks in computer vision field, from challenges like Optical Character Recognition (OCR), Visual Question Answering (VQA), and Referring Expression Comprehension (REC), to more complex, multistep reasoning tasks \cite{fu2024mmecomprehensiveevaluationbenchmark, yu2023mmvetevaluatinglargemultimodal}.

% The mainstream practice in MLLMs employs sequential visual representation, where images are extracted into vision tokens and sent into an LLM. With modal alignment and instruction fine-tuning, modern MLLMs are able to tackle various tasks, ranging from traditional computer vision problems such as Optical Character Recognition (OCR), Visual Question Answering (VQA), and Visual Grounding to complex, multistep reasoniong challenges\cite{fu2024mmecomprehensiveevaluationbenchmark, yu2023mmvetevaluatinglargemultimodal}

% As MLLMs evolve to process higher-resolution images and more sophisticated tasks, the number of tokens grows substantially, increasing computational demands. We concentrate on researching token pruning that preserves model performance, particularly for tasks like Referring Expression Comprehension (REC).

\subsection{Referring Expression Comprehension}
Referring Expression Comprehension (REC), also known as visual grounding~\cite{kazemzadeh-etal-2014-referitgame, qiao2020referring}, aims to localize a specific object in an image described by a referring expression phrased in natural language expressions. As a cross-modal recognition task, REC has seen recent advancements driven by MLLMs~\cite{chen2023shikra, ferret, cogvlm, chen2023minigptv2}. Some of these models have even achieved results comparable to specialized models~\cite{kamath2021mdetr, groundingdino}. These advancements highlight the ability of MLLMs to perform fine-grained object localization and spatial reasoning..
% like Shikra, MiniGPTv2 This task represents a critical benchmark for evaluating the fine-grained visual understanding capabilities of MLLMs. 

\subsection{Token Reduction}
MLLMs typically take in a large amount of visual tokens as the prefix content. For example, LLaVA-1.5~\cite{liu2023improvedllava} encodes images at a resolution of 336 × 336 into 576 tokens and processes images with a greater resolution of 672 × 672 into 2304 tokens. Similarly, MiniGPTv2~\cite{chen2023minigptv2} encodes images at 448 × 448 resolution into 256 tokens by concatenating four adjacent visual tokens. However, the quadratic complexity of Transformers presents a significant challenge when dealing with long input sequences. To address this, several works have attempted to reduce visual tokens by merging them before passing them into LLMs or pruning less important tokens from Transformer layers, thereby lowering computational costs.

PruMerge~\cite{prumerge} leverages the sparse distribution of attention scores between the [CLS] token and visual tokens in ViT~\cite{openaiclip} to measure the importance of each token. It prunes visual tokens with low attention scores and merges them with the selected tokens. On average, LLaVA-1.5 with PruMerge reduces visual tokens to just 6.9\% and cuts 88\% of FLOPs, while maintaining comparable performance in various tasks. SparseVLM~\cite{Zhang2024SparseVLMVT} selects visual-relevant text tokens to measure the significance of visual tokens within the self-attention matrix and prunes the visual tokens accordingly. When reducing tokens from 576 to 64, LLaVA-1.5 with SparseVLM reduces 84\% FLOPs with only a 13.1\% drop in average accuracy across several benchmarks. These works demonstrate that token reduction serves as an effective technique to reduce computational costs while maintaining performance. TRIM~\cite{trim} improves token selection by leveraging the CLIP metric and an Interquartile Range (IQR) scoring function, adaptively selecting important tokens while aggregating information from pruned ones. These works collectively demonstrate that token reduction is an effective technique for reducing computational costs while preserving model performance.

In our study, we focus on how token reduction disrupts the positional relationships between tokens and its impact on the model. We found that directly pruning tokens significantly affects the model’s understanding of spatial relationships between objects, leading to noticeable performance drops in visual grounding tasks like REC.

\section{Preliminaries}
\label{sec:preliminaries}

% figure for sec4 ----------
% Figure for Section 4
\begin{figure*}[h]
  \centering
  \begin{subfigure}{0.48\linewidth}
    \centering
    \includegraphics[height=2in]{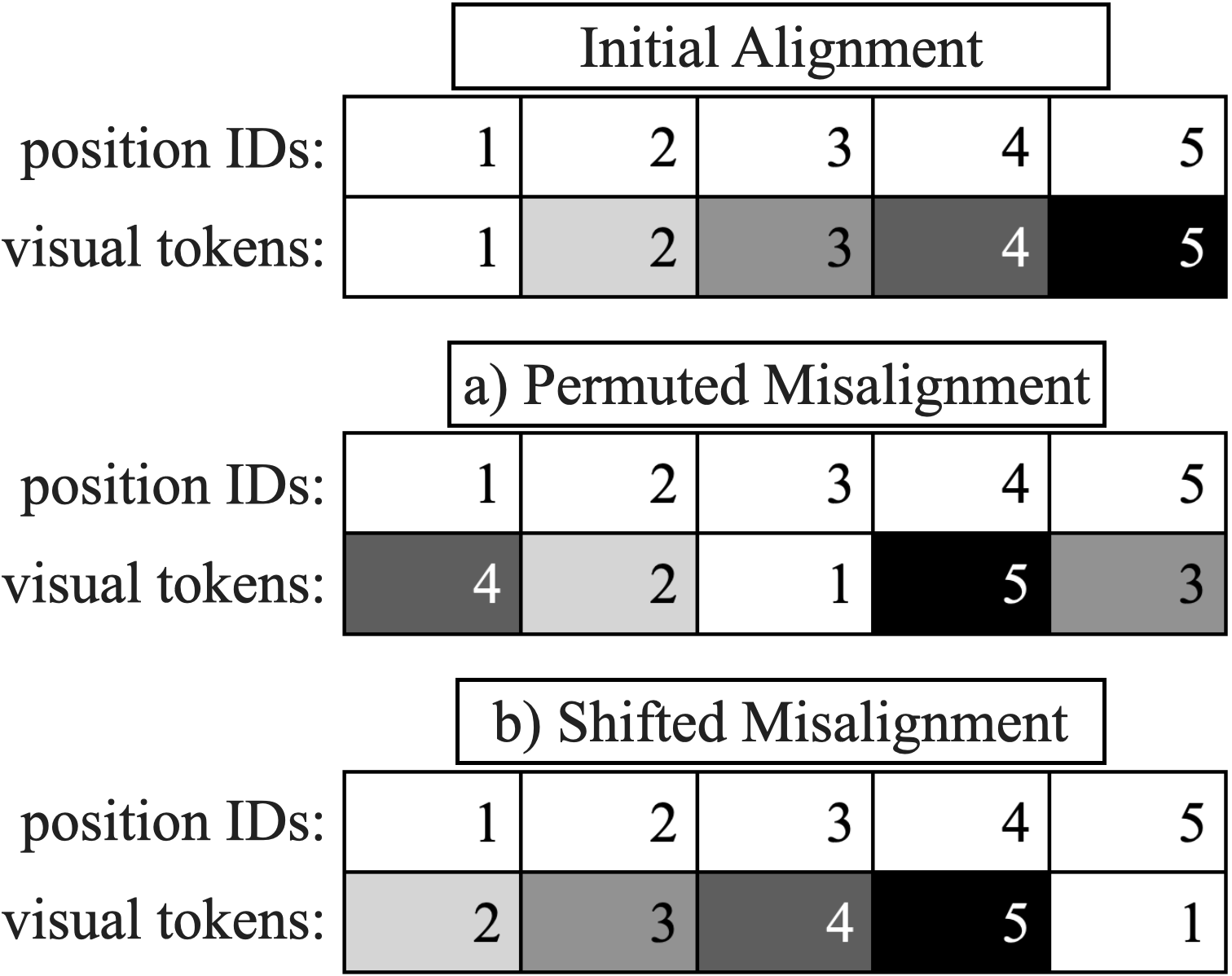}
    \caption{Illustraction of two experiments.}
    \label{fig:observation-demonstration}
  \end{subfigure}
  \hspace{-10pt}
  \begin{subfigure}{0.48\linewidth}
    \centering
    \includegraphics[height=2in]{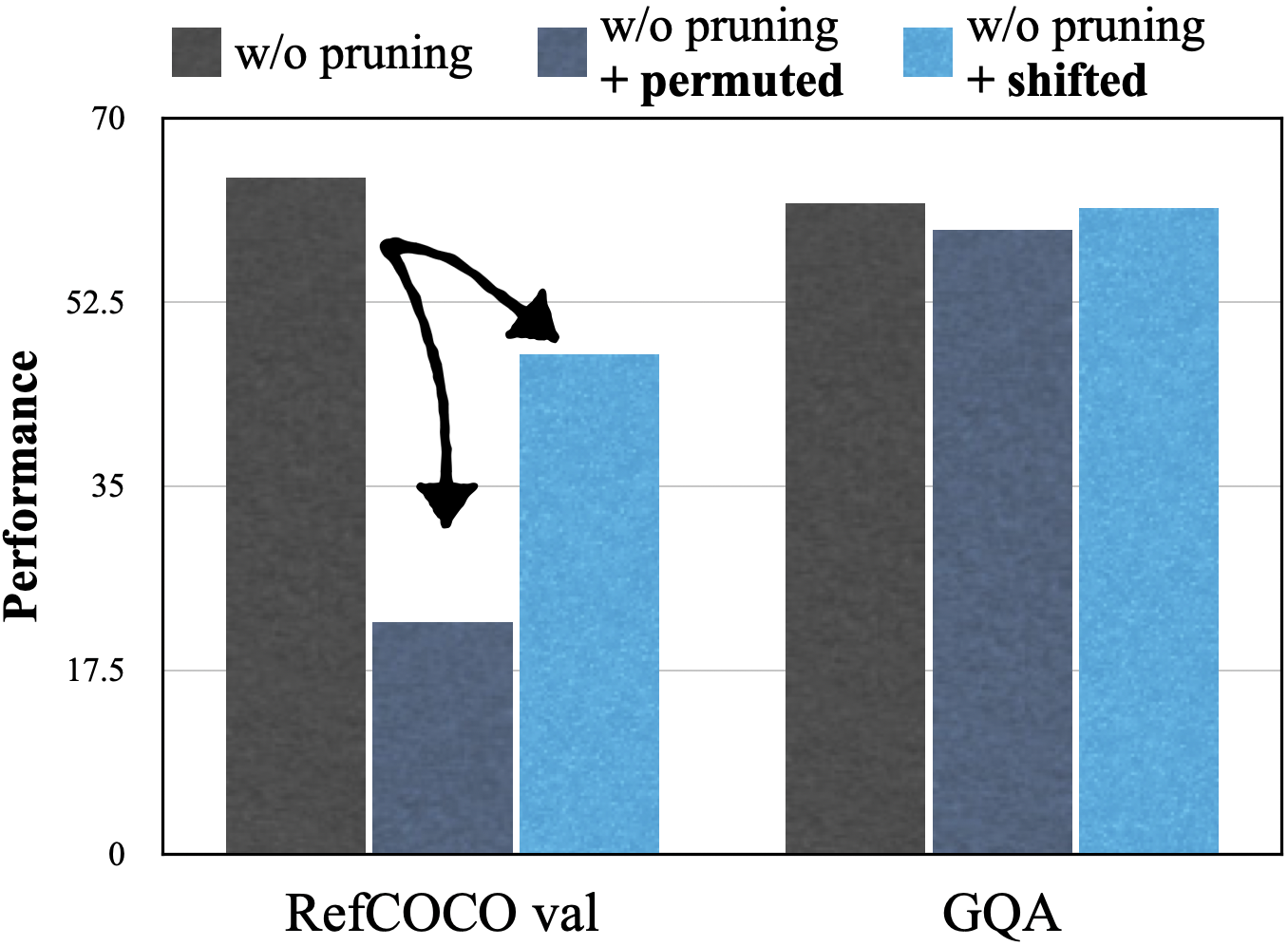}
    \caption{Performance degradation due to position ID misalignment.}
    \label{fig:observation-performance}
  \end{subfigure}
  \caption{\textbf{Visual Token and Position ID Misalignment.} We analyze two types of misalignment caused by pruning and conduct experiments on LLaVA as described in \cref{sec:misalignment}. The results show that while these misalignments have minimal impact on the vision question answering task (GQA), they lead to significant performance degradation in the grounding task (RefCOCO), highlighting a major limitation that must be resolved.}
\label{fig:observation}
  \vspace{-0.5cm}
\end{figure*}

% sec3 started here
During the pruning process, we are given a visual token sequence \( V \in \mathbb{R}^{N \times d} \), a score vector \( S \in \mathbb{R}^{N} \), and a reduction ratio \( r \in (0,1] \). The score vector \( S \) can be computed based on various criteria, such as the similarity between the [CLS] token and the visual tokens, referred to as \textbf{\emph{CLS-visual similarity}} (\cref{eq:cls-visual}), or the similarity between the text token vector \( T \) and the visual tokens \( V \), referred to as \textbf{\emph{text-visual similarity}} (\cref{eq:text-visual}). The reduction ratio \( r \) determines the fraction of tokens to retain, which can be either predefined or adaptively selected.

The number of retained tokens \( k \) is calculated as:
\begin{equation}
    \label{topk}
    k = N \times r
\end{equation}
where \( N \) is the total number of visual tokens.

Next, the top-k tokens are selected by obtaining the indices \( I \) corresponding to the highest values in the score vector \( S \):
\begin{equation}
    \begin{aligned}
    \label{idx}
    I &= topk(S, k) \\
    I &\in \mathbb{Z}^{k}
    \end{aligned}
\end{equation}

Finally, the pruned visual token sequence \( V_{\text{pruned}} \) is generated by selecting the tokens corresponding to the top-k indices:
\begin{equation}
    \label{indicing}
    V_{\text{pruned}} = V[I]
\end{equation}

Through this process, the length of the visual token sequence is reduced from \( N \) to \( r \times N \), improving computational efficiency during inference.
\section{Approach}

\subsection{Our Observation}
\label{sec:observation}
We first observed a significant drop in REC performance during the pruning process while conducting RefCOCO experiments, as shown in \cref{fig:drastic_drop}. The effect was even more pronounced on MiniGPTv2, where we found that pruning as little as $1\%$ of the tokens could lead to a dramatic performance decline, reducing accuracy from $87$ to as low as $18$ (\cref{tab:observation}). To investigate the cause of this, we conducted an empirical study. 

% observation table 1
\begin{table}[h]
  \centering
  \setlength{\tabcolsep}{4.5pt}
  \begin{tabular}{cc|cc}
    \toprule
    \multirow{2}{*}{\textbf{Method}} & \textbf{Tokens} & {RefCOCO} & \multirow{2}{*}{{GQA}}\\
     & \textbf{\%} & val & \\
    \midrule
    \rowcolor{gray!30}w/o pruning & 100\% & 87.35* & 59.6*\\
    \midrule
    CLS-visual & 99\% & 18.58 & 58.77\\
     \bottomrule
  \end{tabular}
  \caption{\textbf{Our Observations.} Pruning even 1\% of tokens in MiniGPTv2 leads to a catastrophic performance drop on RefCOCO, while GQA remains largely unaffected. This suggests that the issue is specific to grounding tasks. An asterisk (*) denotes our reproduced performance.}
  \label{tab:observation}
  \vspace{-0.5cm}
\end{table}

\subsection{Hypothesis}
\label{sec:hypothesis}
We observe that this issue is specific to grounding tasks. For datasets outside of grounding, such as the vision question answering task GQA~\cite{gqa}, previous methods like PruMerge and TRIM~\cite{prumerge, trim} result in less than a 7\% performance drop even when pruning 75\% of tokens. 

Compared to GQA, grounding tasks suffer a larger drop in performance after token pruning (\cref{tab:observation}). We suspect this is because the grounding tasks are more sensitive to position IDs, which are affected during the pruning process. Additionally, the LLM relies on position IDs as the main source of spatial information. To better explain the phenomenon, we have two hypotheses. First, the misalignment between visual tokens and position IDs will severely hurts the performance. Second, the spatial information in ViT is insufficient for the LLM to effectively utilize. To test these hypotheses, we design and conduct experiments on LLaVA.

% linear probing
\begin{table*}[t]
  \centering
  \setlength{\tabcolsep}{4.5pt}
  \begin{tabular}{lcccc}
    \toprule
    \multirow{2}{*}{\textbf{Model}} & \multicolumn{4}{c}{Position Prediction Top-1 Acc.} \\
    \cmidrule(lr){2-5}
     & Layer 1 & Layer 12 & Layer 17 & layer 23 \\
    \midrule
    LLaVA Vision Encoder & 98.37 & 61.92 & 7.30 & 2.61 \\
    \bottomrule
  \end{tabular}
  \caption{\textbf{Spatial Information Retention in Vision Preprocessing of MLLMs.} We train a linear model to predict position IDs, quantifying spatial information retention. Prediction accuracy declines in deeper layers of the LLaVA Vision Encoder, indicating that spatial information is nearly lost by Layer 23, the input vision feature layer for LLaVA’s LLM component. See~\cref{sec:probing} for more details}
\label{tab:probing}
\vspace{-0.5cm}
\end{table*}

\subsubsection{Misalignment between Visual Tokens and Position IDs in LLMs}
\label{sec:misalignment}
Consider a scenario without pruning, where a set of visual tokens \( V = \{v_1, v_2, v_3, v_4, v_5\} \) is obtained from the vision encoder and assigned corresponding position IDs based on their order. These position IDs, denoted as \( P = \{p_1, p_2, p_3, p_4, p_5\} \), are assigned before the tokens are fed into the LLM component of the MLLM. This represents the initial alignment without pruning, as illustrated in~\cref{fig:observation-demonstration}.

However, when pruning occurs, certain visual tokens are removed before position IDs are assigned. As a result, the remaining tokens are mapped to the position IDs based on their updated sequence, rather than preserving their original positional relationships. This disruption introduces misalignment between visual tokens and their position IDs. We observe that token pruning leads to misalignment in two distinct ways, which we analyze in the following sections.

First, token pruning is based on a score vector \( S \) that orders tokens differently from their original sequence (\cref{sec:preliminaries}). For example, if \( S = [4, 2, 1, 5, 3] \), the tokens are rearranged as \( \{v_1, v_4, v_3, v_0, v_2\} \). However, without \emph{GAP}, their position IDs are assigned solely based on the order in which they are input to the LLM, remaining as \( \{p_1, p_2, p_3, p_4, p_5\} \). This results in a permuted misalignment between tokens and their original positional structure, as shown in~\cref{fig:observation-demonstration}.

Second, when pruning removes certain tokens, the remaining tokens are assigned consecutive position IDs starting from \( p_1 \) instead of maintaining their original positions. For instance, if tokens \( v_1 \) and \( v_2 \) are removed, the remaining tokens \( \{v_3, v_4, v_5\} \) are assigned position IDs \( \{p_1, p_2, p_3\} \), shifting their relative positions and causing a shifted misalignment. This effect is also visualized in~\cref{fig:observation-demonstration}.

Since pruning modifies \( V \) but leaves \( P \) unchanged, a misalignment occurs, altering spatial relationships and potentially degrading grounding performance. We propose that both permuted misalignment and shifted misalignment, disrupt the spatial relationships among tokens, altering their relative distances and leading to a drop in grounding performance.

To isolate the effects of these misalignments from the impact of pruning itself, we designed two experiments without removing any visual tokens (see~\cref{fig:observation-demonstration}). As shown in~\cref{fig:observation-performance}, even without losing any visual tokens, the presence of both types of misalignment alone results in performance degradation on the grounding task RefCOCO, highlighting the importance of reducing misalignment to maintain grounding performance.

\subsubsection{Position IDs in the Vision Encoder (ViT)}
\label{sec:probing}
As a preliminary step, the vision encoder assigns position IDs to the image patches before preprocessing it to extract visual tokens. This means that before reaching the LLM, the visual features already contain a structured positional representation. However, since the vision encoder is pretrained on tasks that do not require detailed spatial information, we suspect that the original spatial information might be lost, making it difficult for the LLM to leverage.

Our second hypothesis investigates the role of these vision encoder-generated position IDs and whether they contribute to the LLM’s spatial understanding. Ideally, if the LLM could effectively utilize these original position IDs, it might not need to construct its own. However, if the LLM is unable to leverage them, this suggests that its internally constructed position IDs become the dominant and primary source of spatial information. This, in turn, highlights a fundamental issue: the misalignment caused by pruning becomes even more critical because the LLM lacks access to an alternative source of accurate position encoding. As a result, addressing misalignment is crucial to maintaining grounding performance.

To investigate whether the spatial information from vision encoder can be utilized, we trained a linear model on top of each layer of the vision encoder to predict position IDs, allowing us to quantify how much spatial information is retained at each depth. As shown in \cref{tab:probing}, the model’s ability to predict position IDs decreases in deeper layers, eventually approaching zero. This demonstrates that the original spatial information is progressively lost as the features pass through the encoder. Consequently, by the time these features reach the LLM, the model must rely heavily on its own constructed position IDs to recover spatial structure. This reliance further amplifies the impact of misalignment caused by pruning, as discussed in \cref{sec:misalignment}, \textbf{making accurate position encoding crucial.}

\begin{figure*}[t]
    \centering
    \includegraphics[width=1\linewidth]
    {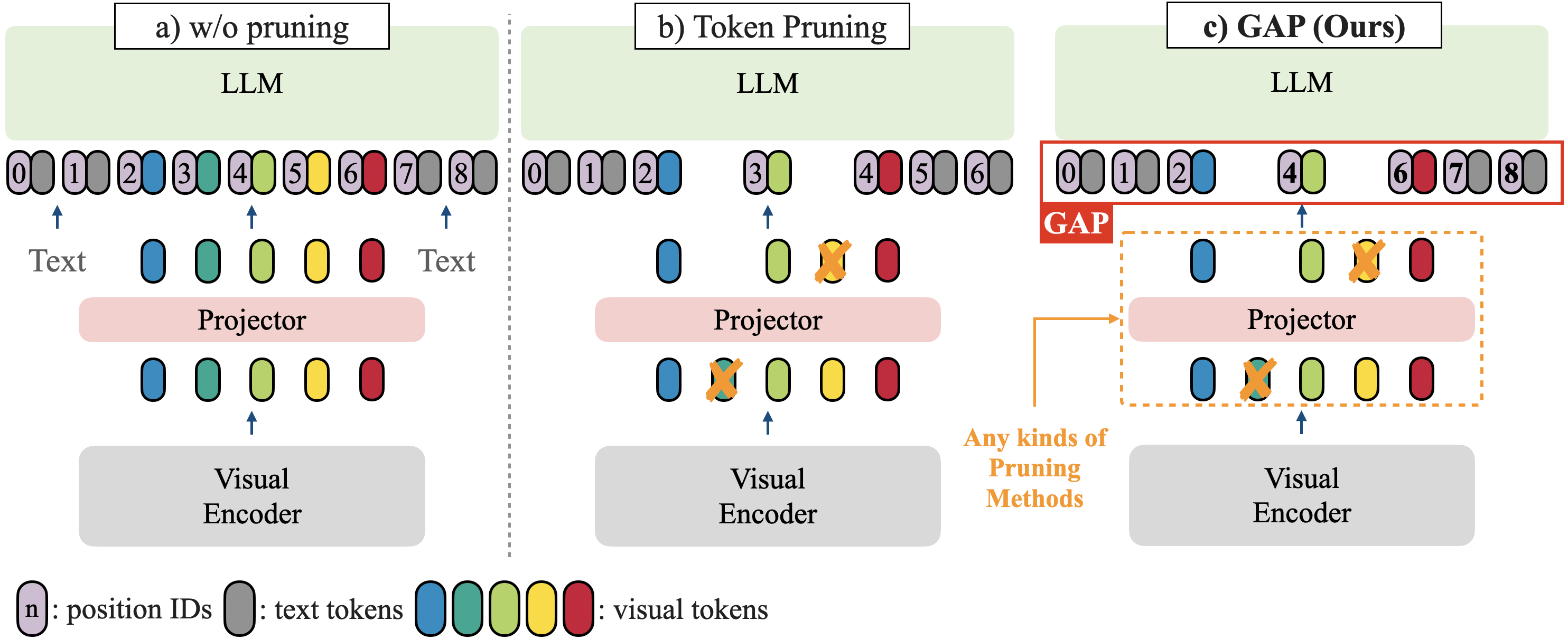}
    \caption{\textbf{Overview of GAP.} (a) The MLLM concatenates visual and text tokens, assigning ordered position IDs. (b) Standard pruning reorders position IDs based on the pruned tokens. (c) GAP discards pruned tokens without incurring additional overhead, while preserving the original position IDs.}
    \label{fig:our_approach}
    \vspace{-0.5cm}
\end{figure*}

\subsection{GAP: Aligning Position IDs}
\label{sec:approach}
Based on our observations, resolving the misalignment between visual token positions and position IDs is crucial for addressing the performance drop observed during pruning. To tackle this issue and enhance performance, we explored several modified approaches for managing position IDs. Ultimately, we propose our solution \emph{GAP} as shown in \cref{fig:our_approach}(c).

\emph{GAP} is an add-on to pruning methods. While pruned tokens are fully discarded, just like in standard token pruning (\cref{fig:our_approach}(b)), the position IDs are retained in their original form, as before pruning (\cref{fig:our_approach}(a)). Specifically, we reconstruct the position IDs as they were prior to pruning, adjust the application of the rotary embedding function to use the modified position IDs, and ensure these updates are applied appropriately during the generation process.

\emph{GAP} prevents the misalignment that leads to performance degradation, as evidenced by the performance improvements in \cref{tab:main_results}. Additionally, it maintains the same computational efficiency as standard token pruning (\cref{tab:ttft}).
\section{Experiments}

\subsection{Evaluation Setup}
\label{sec:setup}

As outlined in~\Cref{sec:preliminaries}, pruning methods for MLLMs can be broadly categorized based on how the score matrix \( S \in \mathbb{R}^N \) is computed, where \( N \) is the total number of visual tokens. We focus on two key approaches: \textbf{CLS-visual similarity pruning} and \textbf{text-visual similarity pruning}.

The \textbf{CLS-visual similarity pruning} approach leverages the [CLS] token, commonly used for classification, and calculate the similarity between the [CLS] token and visual tokens. This method is referred to as \emph{CLS-visual similarity pruning}. In our experiments (see Tables~\ref{tab:main_results} and \ref{tab:cross_method}), we follow the PruMerge \citep{prumerge} setting. The score vector \( S \) is obtained by projecting the [CLS] token into a query \( q_{\text{cls}} \) and the visual tokens into keys \( K \), where \( d_k \) represents the hidden size.

\begin{equation}
    \label{eq:cls-visual}
    S = \text{softmax}\left(\frac{q_{\text{cls}} \cdot K^T}{\sqrt{d_k}}\right)
\end{equation}

The \textbf{text-visual similarity pruning} approach addresses the limitations of CLS-visual pruning by incorporating text information. Instead of relying solely on the [CLS] token, this method computes the similarity between all text tokens and visual tokens, summing over the text dimension. The score vector \( S \) is computed using text queries \( Q_{\text{text}} \) and visual keys \( K_{\text{visual}} \), where \( d_k \) represents the hidden size.

\begin{equation}
    \label{eq:text-visual}
    S = \sum_{\text{text}} \text{softmax}\left( \frac{ Q_{\text{text}} \cdot K_{\text{visual}}^T}{\sqrt{d_k}} \right)
\end{equation}

This method shares a similar principle with SparseVLM \citep{Zhang2024SparseVLMVT}, and has demonstrated strong performance in certain cases.

Additionally, we present two common baseline approaches for comparison in token pruning: \textbf{random pruning} and \textbf{spatial pruning}.

In \textbf{random pruning}, a random score vector \( S \) is generated as:

\begin{equation}
    \label{eq:random}
    S = \mathbf{r}, \quad \mathbf{r} \sim \mathcal{U}(0, 1)^N
\end{equation}
where \( \mathbf{r} \) is a random vector of size \( N \), sampled from a uniform distribution \( \mathcal{U}(0, 1) \).

In \textbf{spatial pruning}, tokens are sampled at regular intervals, determined by the reduction ratio \( r \):

\begin{equation}
\label{eq:spatial}
\begin{split}
V_{\text{pruned}} = \left\{ v_i \mid i = \left\lfloor n \cdot \frac{N}{r} \right\rfloor, \right. \\
\left. n = 0, 1, 2, \ldots \right\}
\end{split}
\end{equation}
where \( V_{\text{pruned}} \) is the set of pruned tokens, and \( v_i \) are the visual tokens at positions \( i \), determined by the reduction ratio \( r \).

We selected models with different configurations of visual encoders, projectors, and LLMs to validate the generalizability of our proposed method, as summarized in Table~\ref{tab:model_comparison}. The models span various sizes, architectures, and pre-training and instruction-tuning datasets, providing a comprehensive evaluation of the effectiveness of \emph{GAP} across diverse settings.

Then, we conducted extensive evaluations of \emph{GAP}, demonstrating its method-agnostic and model-agnostic properties (\cref{sec:main_results}). Our results highlight \emph{GAP}'s effectiveness across different MLLMs, with no additional computational overhead, as confirmed by the inference efficiency comparison (\cref{sec:efficiency}). Furthermore, we evaluate \emph{GAP} across different pruning methods and token reduction ratios (\cref{sec:cross_method}) to assess its generalization. Finally, we verify that \emph{GAP} does not negatively impact VQA performance (\cref{sec:vqa_results}).

\subsection{Main Results}
% fullpipeline results table
\begin{table}[h]
  \centering
  \small
  \setlength{\tabcolsep}{3pt}
  \begin{tabular}{l l c|ccc}
    \toprule
    \multirow{2}{*}{\textbf{Models}} & \multirow{2}{*}{\textbf{Method}} & \textbf{Tokens} & 
    \multicolumn{3}{c}{\textbf{RefCOCO}} \\
    & & ($\%$) & val & testA & testB \\
    \midrule
    \multirow{5}{*}{\textbf{LLaVA-v1.5-7B}} 
    & \cellcolor{gray!30} w/o pruning & \cellcolor{gray!30} 100\% & \cellcolor{gray!30} 56.14 & \cellcolor{gray!30} 64.41 & \cellcolor{gray!30} 47.52 \\
    \cmidrule(lr){2-6}
    & TRIM & 21\% & 3.15 & 3.69 & 3.53 \\
    & + GAP & 21\% & \textbf{17.90} & \textbf{26.30} & \textbf{21.05} \\
    \cmidrule(lr){2-6}
    & PruMerge+ & 25\% & 3.56 & 3.89 & 5.63 \\
    & + GAP & 25\% & \textbf{28.70} & \textbf{30.21} & \textbf{28.38} \\
    \bottomrule
  \end{tabular}
  \caption{\label{tab:fullpipeline_results}\textbf{Results on SOTA methods.} Integrating GAP into both TRIM and PruMerge+ leads to improved performance.}
\end{table}

% model comparison
\begin{table*}[b!]
  \centering
  \small
  \begin{tabular}{l|c|c|c|c|c}
    \toprule
    \textbf{Model} & \textbf{Visual Encoder} & \textbf{Projector} & \textbf{LLM} & \textbf{PT} & \textbf{IT} \\
    \midrule
    \textbf{LLaVA-v1.5-7B} & CLIP-L-336px & MLP & Vicuna-7B & 558K & 665K \\
    \textbf{Llama-LLaVA-NeXT-8B} & CLIP-L-336px, anyres & MLP & Llama3-8B & 558K & 760K \\
    \textbf{LLaVA-v1.6-13B} & CLIP-L-336px, anyres & MLP & Vicuna-13B & 558K & 760K \\
    \textbf{MiniGPTv2} & EVA & Linear* & Llama2-chat 7B & \multicolumn{2}{c}{Three-stage training\dag} \\
    \textbf{Shikra} & CLIP-L & Linear & Vicuna-7B & 600K & 5.5M \\
    \bottomrule
  \end{tabular}
  \caption{\label{tab:model_comparison} \textbf{Model Comparison.} Summary of visual encoders, projectors, large language models (LLM), size of pre-training datasets (PT), and size of instruction tuning datasets (IT) across different models. * indicates a concatenate process before projection. \dag indicates a special three-stage training process; details can be found in the original paper~\cite{chen2023minigptv2}.}
\end{table*}

\label{sec:main_results}
We first apply \emph{GAP} to the state-of-the-art (SOTA) pruning methods, PruMerge and TRIM~\cite{prumerge, trim}, to demonstrate \emph{GAP}'s effectiveness and reveal potential oversights in current pruning strategies. These methods employ advanced ranking and selection mechanisms, adaptive ratio adjustments, and additional merging or pooling techniques. However, regardless of their specific algorithms, they exhibit limitations in grounding tasks. In contrast, \emph{GAP} improves their performance as shown in~\Cref{tab:fullpipeline_results}.

% main results table
\begin{table*}[t]
  \centering
  \small
  \setlength{\tabcolsep}{5pt}
  \begin{tabular}{clc|cccccccc|r}
    \toprule
    \multirow{2}{*}{\textbf{Models}} & \multirow{2}{*}{\textbf{Method}} & \textbf{Tokens} & 
    \multicolumn{3}{c}{\textbf{RefCOCO}} & 
    \multicolumn{3}{c}{\textbf{RefCOCO+}} & 
    \multicolumn{2}{c}{\textbf{RefCOCOg}} & \multirow{2}{*}{\textbf{Avg.}}\\
    & & ($\%$) & val & testA & testB & val & testA & testB & val & testA \\
    \midrule
    \multirow{4}{*}{\makecell{\textbf{LLaVA-v1.5} \\ \textbf{7B}}} & \cellcolor{gray!30}w/o pruning* & \cellcolor{gray!30}$100\%$ & \cellcolor{gray!30}56.14 & \cellcolor{gray!30}64.41 & \cellcolor{gray!30}47.52 & \cellcolor{gray!30}50.04 & \cellcolor{gray!30}59.4 & \cellcolor{gray!30}39.03 & \cellcolor{gray!30}48.61 & \cellcolor{gray!30}48.41 & \cellcolor{gray!30}$100\%$ \\
    & CLS-visual & $50\%$ & 15.34 & 17.08 & 12.88 & 13.28 & 15.26 & 10.96 & 12.89 & 12.90 & 26.8\% \\
    & + GAP & $50\%$ & 51.42 & 59.01 & 44.44 & 44.96 & 54.05 & 37.08 & 45.85 & 45.26 & \textbf{92.5\%} \\
    & $\Delta$ & ~ & \textcolor{customgreen}{+36.08} & \textcolor{customgreen}{+41.93} & \textcolor{customgreen}{+31.56} & \textcolor{customgreen}{+31.68} & \textcolor{customgreen}{+38.79} & \textcolor{customgreen}{+26.12} & \textcolor{customgreen}{+32.96} & \textcolor{customgreen}{+32.36} & \textcolor{customgreen}{\textbf{+ 65.7\%}} \\
    \midrule
    \multirow{4}{*}{\makecell{\textbf{LLaVA-v1.6} \\ \textbf{13B}}} & \cellcolor{gray!30}w/o pruning* & \cellcolor{gray!30}$100\%$ & \cellcolor{gray!30}87.58 & \cellcolor{gray!30}91.24 & \cellcolor{gray!30}80.33 & \cellcolor{gray!30}80.82 & \cellcolor{gray!30}87.91 & \cellcolor{gray!30}70.61 & \cellcolor{gray!30}84.72 & \cellcolor{gray!30}83.33 & \cellcolor{gray!30}$100\%$ \\
    & CLS-visual & $50\%$ & 56.35 & 65.28 & 49.07 & 52.79 & 61.42 & 41.93 & 48.56 & 47.91 & 63.3\% \\
    & + GAP & $50\%$ & 75.50 & 79.88 & 70.15 & 69.65 & 76.86 & 61.08 & 72.78 & 70.32 & \textbf{86.4\%} \\
    & $\Delta$ & ~ & \textcolor{customgreen}{+19.15} & \textcolor{customgreen}{+14.60} & \textcolor{customgreen}{+21.08} & \textcolor{customgreen}{+16.86} & \textcolor{customgreen}{+15.44} & \textcolor{customgreen}{+19.15} & \textcolor{customgreen}{+24.22} & \textcolor{customgreen}{+22.41} &	\textcolor{customgreen}{\textbf{+ 23.1\%}} \\
    \midrule
    \multirow{4}{*}{\makecell{\textbf{Llama3} \\ \textbf{LLaVA-NeXT} \\ \textbf{8B}}} & \cellcolor{gray!30}w/o pruning* & \cellcolor{gray!30}$100\%$ & \cellcolor{gray!30}85.96 & \cellcolor{gray!30}91.18 & \cellcolor{gray!30}76.27 & \cellcolor{gray!30}79.51 & \cellcolor{gray!30}87.91 & \cellcolor{gray!30}68.35 & \cellcolor{gray!30}82.19 & \cellcolor{gray!30}80.62 & \cellcolor{gray!30}$100\%$ \\
    & CLS-visual & $50\%$ & 49.16 & 59.09 & 40.41 & 44.49 & 56.3 & 34.73 & 42.02 & 41.29 & 56.0\%\\
    & + GAP & $50\%$ & 73.48 & 78.95 & 63.12 & 67.48 & 76.32 & 57.35 & 70.72 & 68.63 & \textbf{85.2\%} \\
    & $\Delta$ & ~ & \textcolor{customgreen}{+24.32} & \textcolor{customgreen}{+19.86} & \textcolor{customgreen}{+22.71} & \textcolor{customgreen}{+22.99} & \textcolor{customgreen}{+20.02} & \textcolor{customgreen}{+22.62} & \textcolor{customgreen}{+28.70} & \textcolor{customgreen}{+27.34} &	\textcolor{customgreen}{+ \textbf{29.2\%}} \\
    \midrule
    \multirow{4}{*}{\textbf{MiniGPTv2}} & \cellcolor{gray!30}w/o pruning & \cellcolor{gray!30}$100\%$ & \cellcolor{gray!30}88.69 & \cellcolor{gray!30}91.65 & \cellcolor{gray!30}85.33 & \cellcolor{gray!30}79.97 & \cellcolor{gray!30}85.12 & \cellcolor{gray!30}74.45 & \cellcolor{gray!30}84.44 & \cellcolor{gray!30}84.66 & \cellcolor{gray!30}100$\%$ \\
    & CLS-visual & $50\%$ & 2.73 & 2.07 & 3.80 & 2.61 & 2.07 & 3.80 & 3.61 & 3.68 & 3.7\% \\
    & + GAP & $50\%$ & 68.91 & 74.01 & 67.94 & 61.47 & 67.04 & 58.19 & 68.03 & 67.53 & \textbf{79.0\%} \\
    & $\Delta$ & ~ & \textcolor{customgreen}{+66.18} & \textcolor{customgreen}{+71.94} & \textcolor{customgreen}{+64.14} & \textcolor{customgreen}{+58.86} & \textcolor{customgreen}{+64.97} & \textcolor{customgreen}{+54.39} & \textcolor{customgreen}{+64.42} & \textcolor{customgreen}{+63.85}	& \textcolor{customgreen}{\textbf{+ 75.3\%}} \\
    \midrule
    \multirow{4}{*}{\textbf{Shikra}} & \cellcolor{gray!30}w/o pruning & \cellcolor{gray!30}$100\%$ & \cellcolor{gray!30}87.65 & \cellcolor{gray!30}91.35 & \cellcolor{gray!30}81.62 & \cellcolor{gray!30}82.42 & \cellcolor{gray!30}87.65	& \cellcolor{gray!30}73.38 & \cellcolor{gray!30}81.92 & \cellcolor{gray!30}82.58 & \cellcolor{gray!30}$100\%$ \\
    & CLS-visual & $50\%$ & 9.40 & 10.35 & 8.56 & 7.50 & 7.73 & 7.60 & 7.72 & 8.48 & \textbf{10.1\%} \\
    & + GAP & $50\%$ & 49.33 & 51.63 & 47.61 & 43.85 & 41.03 & 46.40 & 42.06 & 40.74 & \textbf{54.4\%} \\
    & $\Delta$ & ~ & \textcolor{customgreen}{+39.93} & \textcolor{customgreen}{+41.28} & \textcolor{customgreen}{+39.05} & \textcolor{customgreen}{+36.35} & \textcolor{customgreen}{+33.30} & \textcolor{customgreen}{+38.80} & \textcolor{customgreen}{+34.34} & \textcolor{customgreen}{+32.26} &	\textcolor{customgreen}{\textbf{+ 44.3\%}} \\
    \bottomrule
  \end{tabular}
  \caption{\label{tab:main_results}\textbf{Main Results.} The table reports accuracy for different models. The "w/o pruning" row denotes the full-token baseline, while the "CLS-visual" row represents results using the CLS-visual similarity pruning method (50\% token reduction). The "+ GAP" row corresponds to CLS-visual with \emph{GAP} applied, and the $\Delta$ row indicates the accuracy improvement from applying \emph{GAP} to CLS-visual pruning, as calculated in~\cref{eq:delta}. The rightmost column (\textbf{Avg.}) normalizes accuracy relative to the w/o pruning baseline (set as 100\%), allowing a direct comparison of retained performance after pruning. Applying \emph{GAP} results in a performance increase ranging from 23\% to 75\%. An asterisk (*) indicates were reproduced due to the lack of official report. See~\cref{sec:main_results} for more details.}
\end{table*}

\textbf{These methods are only implemented on LLaVA-v1.5}, providing a limited scope for exploration and analysis. Additionally, they utilize adaptive pruning ratios, making it challenging to isolate the direct impact of pruning on grounding ability. To address this, we shift our focus to a more controlled and standardized evaluation setting. 

Specifically, we adopt a widely used pruning strategy by employing \textbf{CLS-visual similarity} as the score vector \( S \) (\cref{eq:cls-visual}), a ranking approach commonly used in prior work \cite{prumerge, Yu2024BalancingPA, Chen2024RecoverableCA}. Furthermore, to ensure consistency across models, we fix the pruning reduction ratio at 0.5. This setup allows us to systematically evaluate the general impact of pruning on grounding ability across different architectures. 

To demonstrate that this issue is a general and global problem, we extend our analysis beyond LLaVA-v1.5 to five different models \cite{chen2023minigptv2, liu2023llava, liu2024llavanext, lmms_eval2024, chen2023shikra}. These models were selected to cover a diverse range of projector types, visual backbones, LLMs, model sizes, training strategies, and training data (detailed in~\Cref{tab:model_comparison}).

We then applied \emph{GAP} to these models. As shown in~\Cref{tab:main_results}, grounding task performance consistently dropped by more than 40\% across all models under pruning, regardless of their architecture. However, \emph{GAP} effectively mitigated this degradation, significantly improving performance on REC datasets. The performance gains achieved by \emph{GAP} range from 23\% to 75\%, as shown in the rightmost column (Avg.), demonstrating its robustness across different architectures. The symbol \( \Delta \) in ~\Cref{tab:main_results} represents the accuracy difference with and without \emph{GAP}, as defined in~\cref{eq:delta}:

\begin{equation}
\label{eq:delta}
{\Delta = Acc_\text{(CLS-visual + GAP)} - Acc_\text{(CLS-visual)}}
\end{equation}

\subsection{Inference Efficiency}
\label{sec:efficiency}
We measured Time-To-First-Token (TTFT) and memory allocation up to the first token during inference on a single 4090 GPU. The evaluation used a batch size of 1 with synchronized timing, averaged over the RefCOCO dataset. An adaptive reduction ratio from PruMerge+ was applied, retaining approximately 25\% of the tokens on average. As shown in Table~\ref{tab:ttft}, \emph{GAP} introduces no additional overhead, ensuring seamless integration without added complexity.

\begin{table}[h]
  \centering
  \setlength{\tabcolsep}{5pt}
  \begin{tabular}{l|cc}
    \toprule
    \textbf{Method}
    & TTFT (ms) & Memory (GB) \\%& Accuracy \\
    \midrule
    \cellcolor{gray!30}\textbf{LLaVA-v1.5} & \cellcolor{gray!30}101.29 & \cellcolor{gray!30}14.70 \\%& \cellcolor{gray!30}61.93 \\
    \cmidrule(lr){1-3}
    Prumerge+ & 52.22 & 13.96 \\%& 51.15\\
    + GAP & 52.46 & 13.96 \\%& 50.51\\
    \bottomrule
  \end{tabular}
\caption{\label{tab:ttft}\textbf{Inference Efficiency.} We measured TTFT and the change in peak memory on the RefCOCO dataset. Demonstrate that applying \emph{GAP} does not introduce additional overhead. See~\cref{sec:efficiency}.}
\vspace{-0.5cm}
\end{table}

\subsection{Generalization Capabilities}
\label{sec:cross_method}
% cross method table
\begin{table}[h]
  \centering
  \setlength{\tabcolsep}{5pt}
  \small
  \begin{tabular}{l|ccc|c}
    \toprule
    \multirow{2}{*}{\textbf{Method}} & 
    \multicolumn{3}{c|}{\textbf{RefCOCO}} & \multirow{2}{*}{\textbf{Avg.}} \\
    & val & testA & testB & \\
    \midrule
    \cellcolor{gray!30}\textbf{LLaVA-v1.5-7b} & \cellcolor{gray!30}56.14 & \cellcolor{gray!30}64.41 & \cellcolor{gray!30}47.52 & \cellcolor{gray!30}100$\%$ \\
    \cmidrule(lr){1-5}
    CLS-visual & 15.34 & 17.08 & 12.88 & 27.0\% \\
    + GAP & \textbf{51.42} & \textbf{59.01} & \textbf{44.44} & \textbf{92.2\%} \\
    \cmidrule(lr){1-5}
    text-visual & 9.23 & 9.83 & 10.42 & 17.9\% \\
    + GAP & \textbf{37.61} & \textbf{42.00} & \textbf{35.37} & \textbf{68.9\%} \\
    \cmidrule(lr){1-5}
    random & 9.25 & 10.57 & 9.42 & 17.6\% \\
    + GAP & \textbf{44.99} & \textbf{52.34} & \textbf{37.76} & \textbf{80.3\%} \\
    \cmidrule(lr){1-5}
    spatial & 7.83 & 9.47 & 8.07 & 15.2\% \\
    + GAP & \textbf{47.08} & \textbf{55.10} & \textbf{40.82} & \textbf{85.1\%} \\
    \midrule
    \cellcolor{gray!30}\textbf{MiniGPTv2} & \cellcolor{gray!30}88.69 & \cellcolor{gray!30}91.65 & \cellcolor{gray!30}85.33 & \cellcolor{gray!30}100$\%$\\
    \cmidrule(lr){1-5}
    CLS-visual & 2.73 & 2.07 & 3.80 & 3.3\% \\
    + GAP & \textbf{68.91} & \textbf{74.01} & \textbf{67.94} & \textbf{79.4\%} \\
    \cmidrule(lr){1-5}
    text-visual & 5.41 & 3.94 & 6.49 & 6.0\% \\
    + GAP & \textbf{77.79} & \textbf{83.80} & \textbf{73.26} & \textbf{88.3\%} \\
    \cmidrule(lr){1-5}
    random & 3.98 & 2.45 & 5.16 & 4.4\% \\
    + GAP & \textbf{79.72} & \textbf{85.46} & \textbf{75.17} & \textbf{90.4\%} \\
    \cmidrule(lr){1-5}
    spatial & 3.03 & 2.12 & 3.61 & 3.3\% \\
    + GAP & \textbf{59.10} & \textbf{59.74} & \textbf{59.48} & \textbf{67.2\%} \\
    \midrule
    \cellcolor{gray!30}\textbf{Shikra} & \cellcolor{gray!30}87.65 & \cellcolor{gray!30}91.35 & \cellcolor{gray!30}81.62 & \cellcolor{gray!30}100$\%$ \\
    \cmidrule(lr){1-5}
    CLS-visual & 9.40 & 10.35 & 8.56 & 10.8\% \\
    + GAP & \textbf{49.33} & \textbf{51.63} & \textbf{47.61} & \textbf{57.0\%} \\
    \cmidrule(lr){1-5}
    text-visual & 7.62 & 7.74 & 6.20 & 8.3\% \\
    + GAP & \textbf{80.51} & \textbf{84.67} & \textbf{76.48} & \textbf{92.7\%} \\
    \cmidrule(lr){1-5}
    random & 9.48 & 8.89 & 9.96 & 10.9\% \\
    + GAP & \textbf{79.01} & \textbf{83.75} & \textbf{71.32} & \textbf{89.7\%} \\
    \cmidrule(lr){1-5}
    spatial & 7.83 & 8.98 & 6.94 & 9.1\% \\
    + GAP & \textbf{80.13} & \textbf{83.08} & \textbf{75.09} & \textbf{91.5\%} \\
    \bottomrule
  \end{tabular}
  \caption{\label{tab:cross_method}\textbf{Cross-method generalization capabilities.} \emph{GAP} generalizes well across various pruning methods and consistently enhances performance across all tested methods. Results are reported for a fixed pruning ratio of 0.5, both with and without \emph{GAP}. The rightmost column normalizes accuracy relative to the w/o pruning baseline. Refer to~\cref{sec:cross_method} for further details.}
  \vspace{-0.5cm}
\end{table}

To further evaluate the generalization capabilities of \emph{GAP}, we conducted experiments across both different pruning methods (\emph{cross-method}) and different token reduction ratios (\emph{cross-ratio}).

\noindent\textbf{Cross-Method Evaluation.}  
We applied \emph{GAP} to multiple pruning methods to assess its effectiveness beyond the primary methods used in the main results. The evaluated methods include:
\begin{itemize}
    \item {CLS-visual pruning}~\cite{prumerge, Yu2024BalancingPA, Chen2024RecoverableCA} (pruning method in~\cref{tab:main_results})
    \item {Text-visual pruning} (\cref{eq:text-visual})~\cite{trim, Yu2024BalancingPA, Chen2024RecoverableCA}
    \item {Random pruning} (\cref{eq:random})
    \item {Spatial pruning} (\cref{eq:spatial})
\end{itemize}

\noindent\textbf{Cross-Ratio Evaluation.} 
Beyond different pruning methods, we also examined the effectiveness of \emph{GAP} across varying pruning reduction ratios. While the primary results were reported with a fixed ratio of 0.5, we extended the evaluation to ratios ranging from 0.2 to 0.8 to determine whether the impact of \emph{GAP} remains consistent across different levels of token pruning. These results further validate the robustness of \emph{GAP} across a broad range of pruning settings.

\begin{figure}[h]
    \centering
    \includegraphics[width=\columnwidth]{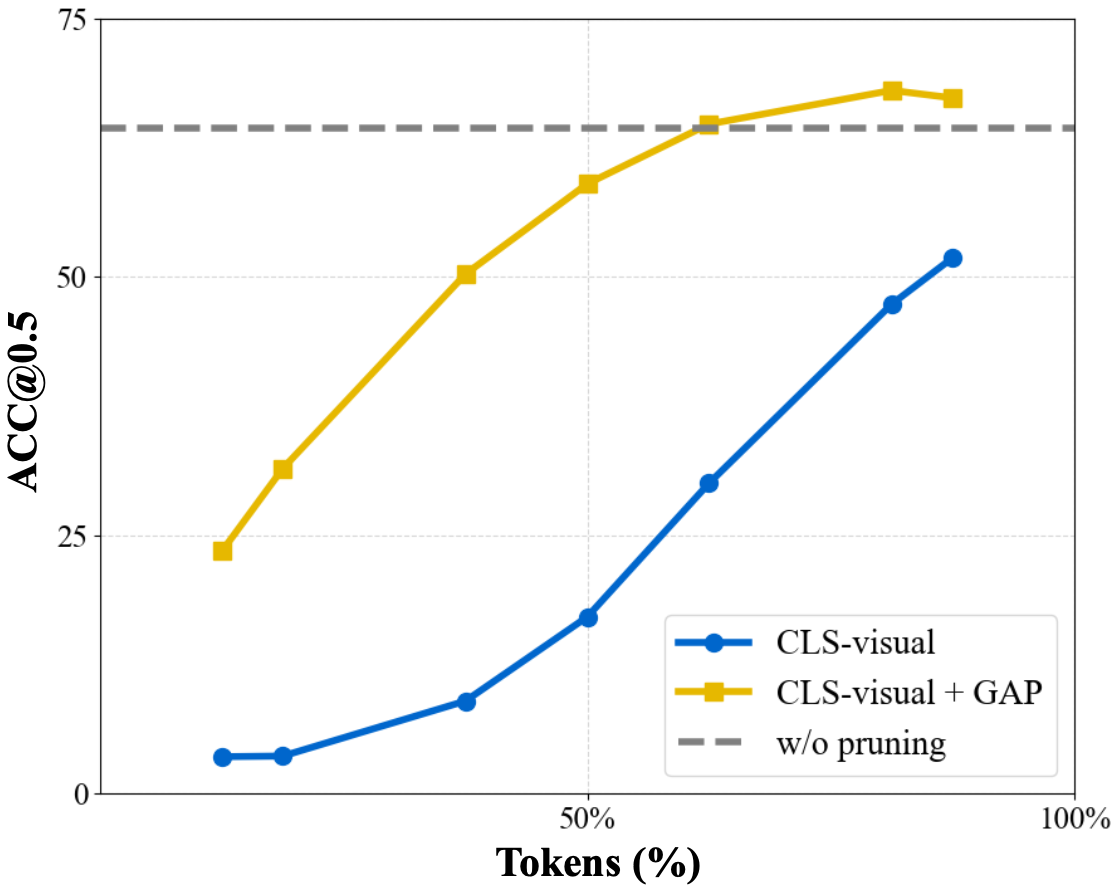}
    \caption{\textbf{Cross-ratio generalization capabilities.} 
    The gray line represents the w/o pruning result with 100\% tokens. This figure shows the accuracy trend across different pruning ratios using the CLS-visual pruning method~\cref{eq:cls-visual}, with and without \emph{GAP}. See~\cref{sec:cross_method} for more details.}
    \label{fig:ratio}
\end{figure}

As shown in~\cref{tab:cross_method} and~\cref{fig:ratio}, the performance drop is consistent across different pruning methods and reduction ratios. More importantly, \emph{GAP} consistently recovers performance across all settings, demonstrating its effectiveness in mitigating pruning-induced degradation regardless of the pruning strategy or token reduction severity. 

One key observation from~\cref{fig:ratio} is that when the pruning ratio is around 75\% to 90\%, the accuracy surpasses the baseline. This aligns with findings in~\cite{proof1, proof2, proof3}, suggesting that a slight reduction in tokens can sometimes enhance performance.

\subsection{Checking Visual Question Answering Results}
\label{sec:vqa_results}
\begin{table}[h]
  \centering
  \small
  \begin{tabular}{l|ccc}
    \toprule
    \textbf{Model} & \textbf{GQA} & \textbf{Vizwiz} & \textbf{OKVQA} \\
    \midrule
    \cellcolor{gray!30}\textbf{LLaVA-v1.5-7B} & \cellcolor{gray!30}61.93 & \cellcolor{gray!30}54.39 & \cellcolor{gray!30}53.44 \\
    \textbf{CLS-visual} & 60.43 & \textbf{54.00} & 53.41 \\
    + GAP & \textbf{61.44} & 53.86 & \textbf{54.49} \\
    \midrule
    \cellcolor{gray!30}\textbf{MiniGPTv2} & \cellcolor{gray!30}59.56 & \cellcolor{gray!30}56.04 & \cellcolor{gray!30}57.99 \\
    \textbf{CLS-visual} & 57.38 & 55.91 & 57.07 \\
    + GAP & \textbf{57.60} & \textbf{56.15} & \textbf{57.65} \\
    \bottomrule
  \end{tabular}
  \caption{\label{tab:vqa_results}\textbf{GAP results on multiple VQA tasks.} GAP demonstrates improvements across various VQA tasks, ensuring it does not negatively impact tasks that do not require grounding. See~\cref{sec:vqa_results} for more details.}
  \vspace{-0.5cm}
\end{table}

Additionally, we expanded our evaluation to include a broader set of datasets for a more comprehensive analysis. We tested \emph{GAP} on GQA, VizWiz, and OKVQA, maintaining a reduction ratio of 0.5. At the beginning, we do not expect GAP to make much difference as the tasks in these datasets are not grounding. Surprisingly, in most cases, our method resulted in slight improvements across these datasets, as illustrated in Table~\ref{tab:vqa_results}. 
Thus, we believe that GAP's improvement depends on the level of spatial relationship understanding required in the datasets.

\section{Conclusions}
\label{sec:conclusions}

As Multimodal Large Language Models (MLLMs) continue to expand, the increasing number of tokens presents a growing challenge for maintaining computational efficiency. Token pruning methods are often employed to address this issue. However, in this work, we identified a significant limitation in existing pruning methods concerning grounding abilities, hypothesizing that the issue stems from the misalignment between visual token positions and positional IDs

To address this, we introduced a simple yet effective method, \emph{GAP}, designed to preserve MLLMs’ grounding ability during token pruning. \emph{GAP} rectifies the misalignment, enabling the model to retain its relational understanding of visual information, which is essential for visual grounding tasks. This effectiveness is demonstrated across five models, including Shikra, MiniGPTv2, and the LLaVA series. Furthermore, \emph{GAP}, as an add-on to existing pruning methods, maintains the same computational efficiency and can be seamlessly integrated, offering a versatile solution that enhances performance without adding complexity.

{
    \small
    \bibliographystyle{ieeenat_fullname}
    \bibliography{main}

\begin{thebibliography}{33}
\providecommand{\natexlab}[1]{#1}
\providecommand{\url}[1]{\texttt{#1}}
\expandafter\ifx\csname urlstyle\endcsname\relax
  \providecommand{\doi}[1]{doi: #1}\else
  \providecommand{\doi}{doi: \begingroup \urlstyle{rm}\Url}\fi

\bibitem[Alec~Radford(2021)]{openaiclip}
Chris Hallacy Aditya Ramesh Gabriel Goh Sandhini Agarwal Girish Sastry Amanda Askell Pamela Mishkin Jack Clark Gretchen Krueger Ilya~Sutskever Alec~Radford, Jong Wook~Kim.
\newblock Learning transferable visual models from natural language supervision.
\newblock \emph{arXiv:2103.00020}, 2021.

\bibitem[Bo~Li* and Liu(2024)]{lmms_eval2024}
Kaichen Zhang* Fanyi Pu* Xinrun Du Yuhao Dong Haotian Liu Yuanhan Zhang Ge Zhang Chunyuan~Li Bo~Li*, Peiyuan~Zhang* and Ziwei Liu.
\newblock Lmms-eval: Accelerating the development of large multimoal models, 2024.

\bibitem[Brown et~al.(2020)Brown, Mann, Ryder, Subbiah, Kaplan, Dhariwal, Neelakantan, Shyam, Sastry, Askell, Agarwal, Herbert-Voss, Krueger, Henighan, Child, Ramesh, Ziegler, Wu, Winter, Hesse, Chen, Sigler, Litwin, Gray, Chess, Clark, Berner, McCandlish, Radford, Sutskever, and Amodei]{brown2020languagemodelsfewshotlearners}
Tom~B. Brown, Benjamin Mann, Nick Ryder, Melanie Subbiah, Jared Kaplan, Prafulla Dhariwal, Arvind Neelakantan, Pranav Shyam, Girish Sastry, Amanda Askell, Sandhini Agarwal, Ariel Herbert-Voss, Gretchen Krueger, Tom Henighan, Rewon Child, Aditya Ramesh, Daniel~M. Ziegler, Jeffrey Wu, Clemens Winter, Christopher Hesse, Mark Chen, Eric Sigler, Mateusz Litwin, Scott Gray, Benjamin Chess, Jack Clark, Christopher Berner, Sam McCandlish, Alec Radford, Ilya Sutskever, and Dario Amodei.
\newblock Language models are few-shot learners, 2020.

\bibitem[Cao et~al.(2023)Cao, Paranjape, and Hajishirzi]{cao-etal-2023-pumer}
Qingqing Cao, Bhargavi Paranjape, and Hannaneh Hajishirzi.
\newblock {P}u{M}er: Pruning and merging tokens for efficient vision language models.
\newblock In \emph{Proceedings of the 61st Annual Meeting of the Association for Computational Linguistics (Volume 1: Long Papers)}, pages 12890--12903, Toronto, Canada, 2023. Association for Computational Linguistics.

\bibitem[Chen et~al.(2023)Chen, Zhang, Zeng, Zhang, Zhu, and Zhao]{chen2023shikra}
Keqin Chen, Zhao Zhang, Weili Zeng, Richong Zhang, Feng Zhu, and Rui Zhao.
\newblock Shikra: Unleashing multimodal llm's referential dialogue magic.
\newblock \emph{arXiv preprint arXiv:2306.15195}, 2023.

\bibitem[Chen et~al.(2021)Chen, Cheng, Gan, Yuan, Zhang, and Wang]{proof3}
Tianlong Chen, Yu Cheng, Zhe Gan, Lu Yuan, Lei Zhang, and Zhangyang Wang.
\newblock Chasing sparsity in vision transformers: An end-to-end exploration.
\newblock \emph{Advances in Neural Information Processing Systems}, 34:\penalty0 19974--19988, 2021.

\bibitem[Chen et~al.(2024)Chen, Xu, Zhang, Liu, Liu, and Liu]{Chen2024RecoverableCA}
Yi Chen, Jian Xu, Xu-Yao Zhang, Wen-Zhuo Liu, Yangyang Liu, and Cheng-Lin Liu.
\newblock Recoverable compression: A multimodal vision token recovery mechanism guided by text information.
\newblock \emph{ArXiv}, abs/2409.01179, 2024.

\bibitem[Devlin et~al.(2019)Devlin, Chang, Lee, and Toutanova]{devlin2019bertpretrainingdeepbidirectional}
Jacob Devlin, Ming-Wei Chang, Kenton Lee, and Kristina Toutanova.
\newblock Bert: Pre-training of deep bidirectional transformers for language understanding, 2019.

\bibitem[Deyao~Zhu(2023)]{zhu2023minigpt}
Xiaoqian Shen Xiang Li Mohamed~Elhoseiny Deyao~Zhu, Jun~Chen.
\newblock Minigpt-4: Enhancing vision-language understanding with advanced large language models.
\newblock \emph{arXiv:2304.10592}, 2023.

\bibitem[Fu et~al.(2024)Fu, Chen, Shen, Qin, Zhang, Lin, Yang, Zheng, Li, Sun, Wu, and Ji]{fu2024mmecomprehensiveevaluationbenchmark}
Chaoyou Fu, Peixian Chen, Yunhang Shen, Yulei Qin, Mengdan Zhang, Xu Lin, Jinrui Yang, Xiawu Zheng, Ke Li, Xing Sun, Yunsheng Wu, and Rongrong Ji.
\newblock Mme: A comprehensive evaluation benchmark for multimodal large language models, 2024.

\bibitem[Gurari et~al.(2018)Gurari, Li, Stangl, Guo, Lin, Grauman, Luo, and Bigham]{vizwiz}
Danna Gurari, Qing Li, Abigale~J. Stangl, Anhong Guo, Chi Lin, Kristen Grauman, Jiebo Luo, and Jeffrey~P. Bigham.
\newblock Vizwiz grand challenge: Answering visual questions from blind people.
\newblock In \emph{2018 IEEE/CVF Conference on Computer Vision and Pattern Recognition}, pages 3608--3617, 2018.

\bibitem[Haoxuan~You(2023)]{ferret}
Zhe Gan Xianzhi Du Bowen Zhang Zirui Wang Liangliang Cao Shih-Fu Chang Yinfei~Yang Haoxuan~You, Haotian~Zhang.
\newblock Ferret: Refer and ground anything anywhere at any granularity.
\newblock \emph{arXiv:2310.07704}, 2023.

\bibitem[Hudson and Manning(2019)]{gqa}
Drew~A. Hudson and Christopher~D. Manning.
\newblock Gqa: A new dataset for real-world visual reasoning and compositional question answering.
\newblock In \emph{2019 IEEE/CVF Conference on Computer Vision and Pattern Recognition (CVPR)}, pages 6693--6702, 2019.

\bibitem[Jun~Chen and Elhoseiny(2023)]{chen2023minigptv2}
Xiaoqian Shen Xiang Li Zechun Liu Pengchuan Zhang Raghuraman Krishnamoorthi Vikas Chandra Yunyang~Xiong Jun~Chen, Deyao~Zhu and Mohamed Elhoseiny.
\newblock Minigpt-v2: Large language model as a unified interface for vision-language multi-task learning.
\newblock \emph{arXiv:2310.09478}, 2023.

\bibitem[Kamath et~al.(2021)Kamath, Singh, LeCun, Misra, Synnaeve, and Carion]{kamath2021mdetr}
Aishwarya Kamath, Mannat Singh, Yann LeCun, Ishan Misra, Gabriel Synnaeve, and Nicolas Carion.
\newblock Mdetr--modulated detection for end-to-end multi-modal understanding.
\newblock \emph{arXiv preprint arXiv:2104.12763}, 2021.

\bibitem[Kazemzadeh et~al.(2014)Kazemzadeh, Ordonez, Matten, and Berg]{kazemzadeh-etal-2014-referitgame}
Sahar Kazemzadeh, Vicente Ordonez, Mark Matten, and Tamara Berg.
\newblock {R}efer{I}t{G}ame: Referring to objects in photographs of natural scenes.
\newblock In \emph{Proceedings of the 2014 Conference on Empirical Methods in Natural Language Processing ({EMNLP})}, pages 787--798, Doha, Qatar, 2014. Association for Computational Linguistics.

\bibitem[Li et~al.(2024)Li, Zhang, Guo, Zhang, Li, Zhang, Zhang, Li, Liu, and Li]{llavaonevision}
Bo Li, Yuanhan Zhang, Dong Guo, Renrui Zhang, Feng Li, Hao Zhang, Kaichen Zhang, Yanwei Li, Ziwei Liu, and Chunyuan Li.
\newblock Llava-onevision: Easy visual task transfer.
\newblock \emph{arXiv preprint arXiv:2408.03326}, 2024.

\bibitem[Liu et~al.(2023{\natexlab{a}})Liu, Li, Li, and Lee]{liu2023improvedllava}
Haotian Liu, Chunyuan Li, Yuheng Li, and Yong~Jae Lee.
\newblock Improved baselines with visual instruction tuning, 2023{\natexlab{a}}.

\bibitem[Liu et~al.(2023{\natexlab{b}})Liu, Li, Wu, and Lee]{liu2023llava}
Haotian Liu, Chunyuan Li, Qingyang Wu, and Yong~Jae Lee.
\newblock Visual instruction tuning.
\newblock In \emph{NeurIPS}, 2023{\natexlab{b}}.

\bibitem[Liu et~al.(2024)Liu, Li, Li, Li, Zhang, Shen, and Lee]{liu2024llavanext}
Haotian Liu, Chunyuan Li, Yuheng Li, Bo Li, Yuanhan Zhang, Sheng Shen, and Yong~Jae Lee.
\newblock Llava-next: Improved reasoning, ocr, and world knowledge, 2024.

\bibitem[Marino et~al.(2019)Marino, Rastegari, Farhadi, and Mottaghi]{okvqa}
Kenneth Marino, Mohammad Rastegari, Ali Farhadi, and Roozbeh Mottaghi.
\newblock Ok-vqa: A visual question answering benchmark requiring external knowledge.
\newblock In \emph{2019 IEEE/CVF Conference on Computer Vision and Pattern Recognition (CVPR)}, pages 3190--3199, 2019.

\bibitem[Shang et~al.(2024)Shang, Cai, Xu, Lee, and Yan]{prumerge}
Yuzhang Shang, Mu Cai, Bingxin Xu, Yong~Jae Lee, and Yan Yan.
\newblock Llava-prumerge: Adaptive token reduction for efficient large multimodal models.
\newblock \emph{arXiv preprint arXiv:2403.15388}, 2024.

\bibitem[Shilong~Liu(2023)]{groundingdino}
Tianhe Ren Feng Li Hao Zhang Jie Yang Qing Jiang Chunyuan Li Jianwei Yang Hang Su Jun Zhu Lei~Zhang Shilong~Liu, Zhaoyang~Zeng.
\newblock Grounding dino: Marrying dino with grounded pre-training for open-set object detection.
\newblock \emph{arXiv:2303.05499}, 2023.

\bibitem[Song et~al.(2024)Song, Wang, Chen, Wang, Guan, and Wang]{trim}
Dingjie Song, Wenjun Wang, Shunian Chen, Xidong Wang, Michael Guan, and Benyou Wang.
\newblock Less is more: A simple yet effective token reduction method for efficient multi-modal llms.
\newblock \emph{arXiv preprint arXiv:2409.10994}, 2024.

\bibitem[Wang et~al.(2023)Wang, Chen, Zhou, Zhu, Liang, Shan, Liu, Xu, Yang, and Qin]{proof2}
Zekun Wang, Jingchang Chen, Wangchunshu Zhou, Haichao Zhu, Jiafeng Liang, Liping Shan, Ming Liu, Dongliang Xu, Qing Yang, and Bing Qin.
\newblock Smarttrim: Adaptive tokens and attention pruning for efficient vision-language models.
\newblock \emph{arXiv preprint arXiv:2305.15033}, 2023.

\bibitem[Weihan~Wang(2023)]{cogvlm}
Wenmeng Yu Wenyi Hong Ji Qi Yan Wang Junhui Ji Zhuoyi Yang Lei Zhao Xixuan Song Jiazheng Xu Bin Xu Juanzi Li Yuxiao Dong Ming Ding Jie~Tang Weihan~Wang, Qingsong~Lv.
\newblock Cogvlm: Visual expert for pretrained language models.
\newblock \emph{arXiv:2311.03079}, 2023.

\bibitem[Xiaohua~Zhai(2023)]{siglip}
Alexander Kolesnikov Lucas~Beyer Xiaohua~Zhai, Basil~Mustafa.
\newblock Sigmoid loss for language image pre-training.
\newblock \emph{arXiv:2303.15343}, 2023.

\bibitem[Yanyuan~Qiao(2020)]{qiao2020referring}
Qi~Wu Yanyuan~Qiao, Chaorui~Deng.
\newblock Referring expression comprehension: A survey of methods and datasets.
\newblock \emph{IEEE TMM}, 2020.
\newblock arXiv:2007.09554.

\bibitem[Yu et~al.(2024)Yu, Chen, and Xu]{Yu2024BalancingPA}
Gaotong Yu, Yi Chen, and Jian Xu.
\newblock Balancing performance and efficiency: A multimodal large language model pruning method based image text interaction.
\newblock \emph{ArXiv}, abs/2409.01162, 2024.

\bibitem[Yu et~al.(2023)Yu, Yang, Li, Wang, Lin, Liu, Wang, and Wang]{yu2023mmvetevaluatinglargemultimodal}
Weihao Yu, Zhengyuan Yang, Linjie Li, Jianfeng Wang, Kevin Lin, Zicheng Liu, Xinchao Wang, and Lijuan Wang.
\newblock Mm-vet: Evaluating large multimodal models for integrated capabilities, 2023.

\bibitem[Yuxin~Fang(2022)]{eva}
Binhui Xie Quan Sun Ledell Wu Xinggang Wang Tiejun Huang Xinlong Wang Yue~Cao Yuxin~Fang, Wen~Wang.
\newblock Eva: Exploring the limits of masked visual representation learning at scale.
\newblock \emph{arXiv:2211.07636}, 2022.

\bibitem[Zhang et~al.(2024{\natexlab{a}})Zhang, Cheng, Lu, Zhuo, Wang, Cao, Guo, She, and Zhang]{proof1}
Qizhe Zhang, Aosong Cheng, Ming Lu, Zhiyong Zhuo, Minqi Wang, Jiajun Cao, Shaobo Guo, Qi She, and Shanghang Zhang.
\newblock [cls] attention is all you need for training-free visual token pruning: Make vlm inference faster.
\newblock \emph{arXiv preprint arXiv:2412.01818}, 2024{\natexlab{a}}.

\bibitem[Zhang et~al.(2024{\natexlab{b}})Zhang, Fan, Ma, Zheng, Huang, Cheng, Gudovskiy, Okuno, Nakata, Keutzer, and Zhang]{Zhang2024SparseVLMVT}
Yuan Zhang, Chun-Kai Fan, Junpeng Ma, Wenzhao Zheng, Tao Huang, Kuan Cheng, Denis~A Gudovskiy, Tomoyuki Okuno, Yohei Nakata, Kurt Keutzer, and Shanghang Zhang.
\newblock Sparsevlm: Visual token sparsification for efficient vision-language model inference.
\newblock 2024{\natexlab{b}}.

\end{thebibliography}
}

\end{document}